\documentclass[10pt,twocolumn,letterpaper]{article}

\usepackage{cvpr}
\usepackage{times}
\usepackage{epsfig}
\usepackage{graphicx}
\usepackage{amsmath}
\usepackage{amssymb}
\usepackage{algorithm}
\usepackage{algpseudocode}

\usepackage[breaklinks=true,bookmarks=false]{hyperref}

\cvprfinalcopy 

\def\cvprPaperID{4159} 

\begin{document}
%
\title{Efficient Multi-Person Pose Estimation with Provable Guarantees}
\author{
Shaofei Wang\\
Beijing A\&E Technologies \\
Beijing, China\\
\texttt{sfwang0928@gmail.com} \\
\and 
Konrad Paul Kording\\
University of Pennsylvania\\
Philadelphia, PA \\
\texttt{koerding@gmail.com} 
\and 
Julian Yarkony\\
Experian Data Labs \\
San Diego, CA\\
\texttt{julian.e.yarkony@experian.com} 
}
\maketitle
%
\begin{abstract}
Multi-person pose estimation (MPPE) in natural images is key to the meaningful use of visual data in many fields including movement science, security, and rehabilitation. In this paper we tackle MPPE with a bottom-up approach, starting with candidate detections of body parts from a convolutional neural network (CNN) and grouping them into people. We formulate the grouping of body part detections into people as a minimum-weight set packing (MWSP) problem where the set of potential people is the power set of body part detections. We model the quality of a hypothesis of a person which is a set in the MWSP by an augmented tree-structured Markov random field where variables correspond to body-parts and their state-spaces correspond to the power set of the detections for that part. 

We describe a novel algorithm that combines efficiency with provable bounds on this MWSP problem.  We employ an implicit column generation strategy where the pricing problem is formulated as a dynamic program. To efficiently solve this dynamic program we exploit the problem structure utilizing a nested Bender's decomposition (NBD) exact inference strategy which we speed up by recycling Bender's rows between calls to the pricing problem. 

We test our approach on the MPII-Multiperson dataset, showing that our approach obtains comparable results with the state-of-the-art algorithm for joint node labeling and grouping problems, and that NBD achieves considerable speed-ups relative to a naive dynamic programming approach. Typical algorithms that solve joint node labeling and grouping problems use heuristics and thus can not obtain proofs of optimality. Our approach, in contrast, proves that for over 99 percent of problem instances we find the globally optimal solution and otherwise provide upper/lower bounds. 
\end{abstract}
%
\section{Introduction}
We study the problem of multi-person pose estimation (MPPE) \cite{mpiiBenchmark} which we model as the problem of selecting a subset of non-overlapping proposals of people supported by image evidence and a prior model.  This formulation of MPPE corresponds to a minimum weight set packing (MWSP)  \cite{karp} problem where elements correspond to detections of body parts and sets (referred to as poses) correspond to subsets of those body parts detections.  Poses are associated with real valued costs based on occurrence and co-occurrence probabilities of detections in a pose as defined by a deep neural network \cite{hinton,krizhevsky2012imagenet,wu2015deep,baldi2014searching} and (augmented) tree structured part/spring model \cite{deva1,deva2,deva3} respectively. This fully specifies MPPE as an optimization problem.  

Since the set of poses grows exponentially in the number of detections we employ an implicit column generation (ICG) strategy \cite{barnprice,cuttingstock,gilmore1965multistage} for inference in MWSP.  We exploit the augmented tree structure of the cost of a pose to frame pricing as a dynamic program \cite{dynProgBook} where variables correspond to body parts and the state of a given variable corresponds to a subset of the detections for that part. Since the state-space of each variable is enormous (power set of detections of a given part), we introduce a tool from operations research called the nested Benders decomposition (NBD) \cite{benders1962partitioning,birge1985decomposition} which avoids considering the vector product of the state-spaces of adjacent variables in the tree. NBD has been used for a variety of applications including: agricultural planning, factory production planning, and stock portfolio optimization.  Our NBD formulation is guaranteed to achieve exact inference in the pricing problem and in practice is orders of magnitude faster than naive dynamic programming.  NBD exploits the fact that pricing problems are similar across iterations of ICG by hot starting optimization in a given iteration with Benders rows produced in previous iterations.  
%
%
  The combination of ICG with NBD promises efficient and provably optimal solutions without having to enumerate the vector product of state-spaces.
\subsection{Related Work}
Our work is closely related to the sub-graph multi-cut integer linear programming (ILP) formulation of \cite{deepcut1,deepcut2,NL-LMP} which we refer to as MC for shorthand.  MC models the problem of MPPE as partitioning detections into fourteen body parts (plus false positive) and clustering those detections into poses.  Clustering is done according to the correlation clustering \cite{corclustorig} criteria with costs parameterized by the part associated with the detection. This formulation is notable as it models non-max suppression by allowing poses to be associated with multiple detections of a given body part.  However, the optimization problem of MC is often too hard to solve and is thus attacked with heuristic methods. 

Motivated by the difficulty of inference in MC, the work of \cite{wang2017multi} introduces an alternative model called the two tier formulation (TTF).  TTF truncates the model of MC in such a way to achieve fast inference using ICG.  The TTF model with ICG inference outperforms MC with regards to finding difficult-to-localize parts such as ankle and wrist, and also provides a marginal improvement in overall accuracy.  Furthermore TTF provides some additional capacities beyond that of MC such as a prior on the number of people in an image.  In \cite{wang2017exploiting} inference in the TTF model is attacked using a non-nested Benders decomposition \cite{geoffrion1974multicommodity,benders1962partitioning} though inference is demonstrated to not be as fast as in the ICG strategy. While the TTF model works well in practice, it does not optimize the full cost but privileges a single exemplar detection for each body part to model inter-part co-association costs of a pose, ignoring all inter-part co-association costs between non-exemplar detections.  Furthermore TTF requires detections to be associated with parts in advance of optimization.  
\subsection{Outline}
Our paper is structured as follows. In Section \ref{myModel} we introduce a novel MWSP formulation for MPPE and address inference with ICG.  In Section \ref{bendPricing} we introduce our NBD approach to solving the pricing problem of ICG. In Section \ref{exper} we present experiments on the MPII-Multiperson validation set.  Finally we conclude in Section \ref{conc}.  We provide additional derivations in the appendix.
\section{Our MWSP Formulation of MPPE}
\label{myModel}
We now formulate MPPE as MWSP.  
We use $\mathcal{R}$,$\mathcal{D}$ to denote the sets of human body parts and the sets of body part detections, which we index with $r$,$d$ respectively.  We describe a surjection of detections to parts using $R\in \{\mathcal{R}\}^{|\mathcal{D}|}$ where $R_{d}$ indicates the part associated with detection $d$.  
Each detection is associated with a single part \textit{prior to MWSP optimization}.  For short hand we use $\mathcal{D}^r$ to denote the set of detections associated with part $r$.
We use $\mathcal{P}$ to denote the set of all possible poses which we index by $p$. We associate the members of $\mathcal{P}$ with detections using $P \in \{0,1\}^{|\mathcal{D}|\times |\mathcal{P}|}$ which we index by $d,p$ where $P_{dp}=1$ indicates that detection $d$ is in pose $p$. This allows us to formulate MPPE as a search for low cost, non-overlapping sets. 

We model human poses with an augmented  tree structure over the set of parts described using matrix $T \in \{0,1\}^{|\mathcal{R}|\times |\mathcal{R}|}$.  We index $T$ with $r_1,r_2$ where $T_{r_1r_2}=1$ if and only if part $r_2$ is a child of part $r_1$ in the augmented tree.  We use fourteen body parts (head, neck, and the left/right of shoulder, elbow, wrist, hip, knee, ankle), as standardized by MPII dataset.  We form an augmented tree over these parts defined by a typical tree structure over all parts excluding the neck, followed by connecting the neck to each of the other thirteen parts; this model design is based on the observation that in real images people's necks are rarely occluded, thus having connections from neck to all other body parts can handle cases where other body parts are occluded while keeping the model relatively simple. For short hand we use $T_0$ to denote the root of the tree which is a part other than the neck selected arbitrarily.

The cost of a pose is defined using terms $\theta^1 \in \mathbb{R}^{|\mathcal{D}|}$, and $\theta^2 \in \mathbb{R}^{|\mathcal{D}|\times |\mathcal{D}|}$ which we index with $d,$ and $d_1,d_2$ respectively.  We refer to the $\theta^1,\theta^2$ terms as unary and pairwise respectively. We use $\theta^1_{d}$ to indicate the cost of including detection $d$ in a pose.   Similarly we use $\theta^2_{d_1d_2}$ to indicate the cost of including detections $d_1,d_2$ in a common pose.  The augmented tree structure is respected with regards to these costs; thus $\theta^2_{d_1,d_2}$ can only be non-zero if $R_{d_1}=R_{d_2}$ or if $R_{d_2}$ is a child of $R_{d_1}$.  
We model a prior on the number of poses in the image using $\theta^0\in \mathbb{R}$ which is a constant offset added to the cost of each pose.  We define the mapping of poses to costs using $\Gamma \in \mathbb{R}^{|\mathcal{P}|}$ which we index with $p$ where $\Gamma_p$ is the cost associated with pose $p$ which is defined formally below.  
\begin{align}
\Gamma_p=\theta^0+\sum_{d \in \mathcal{D}}\theta^1_{d}P_{dp}+\sum_{(d_1,d_2) \in \mathcal{D}}\theta^2_{d_1d_2}P_{d_1p}P_{d_2p}
\label{eqn:pose-cost}
\end{align}
We have thus fully defined the cost of a single pose. In the next section we formulate the MWSP problem using the costs of poses.
\subsection{ILP/ LP Relaxation Formulation}
We frame the search for the lowest cost set of non-overlapping poses as an integer linear program (ILP).  We use $\gamma \in \{0,1\}^{|\mathcal{P}|}$ to define a selection of poses where $\gamma_p=1$ if and only if pose $p$ is selected.   We write the constraint that selected poses do not overlap as $P\gamma \leq 1$.  We express the ILP along with the corresponding primal/dual linear program (LP) relaxations as follows using Lagrange multipliers $\lambda \in \mathbb{R}_{0+}^{|\mathcal{D}|}$ which we index by $d$.  
\begin{align}
\label{ilpLpOrig}
\min_{\substack{\gamma \in  \{0,1\} \\ P\gamma \leq 1}}\Gamma^{\top}\gamma
 \geq \min_{\substack{\gamma \geq 0 \\ P\gamma \leq 1}}\Gamma^{\top}\gamma
 =\max_{\substack{\lambda \geq 0 \\ \Gamma +P^{\top}\lambda \geq 0}}-1^{\top}\lambda
\end{align}
In Appendix \ref{dualBound} 
we study dual-optimal bounds \cite{ben2006dual,HPlanarCC} on $\lambda$ that when applied can accelerate inference without loosening the LP relaxation.  
\subsection{Inference via Implicit Column Generation}
Given that $\mathcal{P}$ grows exponentially in the number of detections we can not explicitly consider it during MWSP optimization. We thus construct a sufficient subset $\hat{\mathcal{P}} \subset \mathcal{P}$ using ICG so as to solve exactly the dual problem in Eq \ref{ilpLpOrig}. ICG consists of two alternating optimizations  referred to as the   restricted master problem (RMP) and the pricing problem respectively.
\begin{itemize}
\item
  RMP:  We solve the  dual optimization over the set $\hat{\mathcal{P}}$ providing dual variables $\lambda$.  
    We write dual optimization of the RMP below.
  \begin{align}
\max_{\substack{\lambda \geq 0 \\ \Gamma_p +\sum_{d \in \mathcal{D}}P_{dp}\lambda_d \geq 0 \quad \forall p \in \hat{\mathcal{P}}}}-1^{\top}\lambda
  \end{align}
  \item Pricing:  Using $\lambda$ we identify a subset of the most violated constraints corresponding to members of $\mathcal{P}$ and add it to $\hat{\mathcal{P}}$.  This subset includes the most violated constraint over all $\mathcal{P}$. When no violated constraints exist ICG terminates. The slack in the dual constraint corresponding to a given primal variable is referred to as its reduced cost, thus pricing identifies the lowest reduced cost terms in the primal.
  
  During the pricing step we iterate through the power set of neck detections and compute the lowest reduced cost pose containing exactly those neck detections.  We index the power set of neck detections with $\breve{\mathcal{D}}$ and use $p \leftrightarrow  \breve{\mathcal{D}}$ to indicate that the neck detections in $p$ are exactly those in  $\breve{\mathcal{D}}$.  We write pricing below for an arbitrary  subset of the neck detections $\breve{\mathcal{D}}$. 
\begin{align}
\label{PricingEq}
\min_{\substack{p \in \mathcal{P}\\ p \leftrightarrow \breve{\mathcal{D}} }} \Gamma_p+\sum_{d\in \mathcal{D}}\lambda_dP_{dp}
\end{align}
 Note that when conditioned on a specific set of neck detections, the pairwise costs from these neck detections to all other detections can be added to unary costs of the other detections.  Thus the augmented-tree structure becomes a typical tree structure, and exact inference can be done via dynamic programming.
  \end{itemize}
We write ICG in Alg \ref{dualSolve} and consider the pricing problem which is a dynamic program in Section~\ref{basicPricing}.  At termination of Alg \ref{dualSolve} we solve MWSP over $\hat{\mathcal{P}}$ using an ILP solver.  In practice we find that the LP relaxation provides an integral solution at termination for over 99\% of cases.  If needed we could employ branch and price~\cite{barnprice} or tighten the bound with odd set inequalities 
while preserving the structure of the pricing problem~\cite{wang2017tracking}.   %
\begin{figure}
 \begin{algorithm}[H]
 \caption{Implicit Column Generation Algorithm}
\begin{algorithmic} 
\State $\hat{\mathcal{P}} \leftarrow \{ \}$
\Repeat
\State did\_add$\leftarrow 0$
\State $\lambda \leftarrow \max_{\substack{\lambda \geq 0 \\ \Gamma_p +\sum_{d \in \mathcal{D}}P_{dp}\lambda_d \geq 0 \quad \forall p \in \hat{\mathcal{P}}}}-1^{\top}\lambda$
\For{$\breve{\mathcal{D}} \subseteq \mathcal{D}^{\mbox{neck}}$}
\State $p \leftarrow \min_{\substack{p \in \mathcal{P}\\p \leftrightarrow \breve{\mathcal{D}}}} \Gamma_p+\sum_{d\in \mathcal{D}} P_{dp}\lambda_d $
\If{$\Gamma_p+\sum_{d\in \mathcal{D}} P_{dp}\lambda_d < 0$}
\State $\hat{\mathcal{P}} \leftarrow [\hat{\mathcal{P}} \cup p] $
\State did\_add$\leftarrow 1$
\EndIf
\EndFor
 \Until{ did\_add=0  }
\end{algorithmic}
\label{dualSolve}
  \end{algorithm}
  \caption{Implicit column generation procedure.  We iteratively solve the RMP followed by pricing.  During pricing we compute one pose associated with each member of the power set of neck detections.  We add the pose to nascent set $\hat{\mathcal{P}}$ if and only if it corresponds to a violated constraint in the dual.  We terminate when no pose in $\mathcal{P}$ corresponds to a violated dual constraint. 
  }
\end{figure}
\subsection{Anytime Lower Bounds}
We compute an anytime lower bound on our objective by adding a term based on the columns produced to the objective of the  RMP. Recall that each detection can be assigned to at most one body part. Thus we can rely on the proof that the LP for MWSP can be bounded by  RMP objective plus the lowest reduced cost  times the cardinality of the set of elements \cite{desrosiers2005primer,wang2017tracking} (if a negative reduced cost term exists).  We compute this lower bound below given any non-negative $\lambda$ provided by the RMP as follows.
\begin{align}
\label{LBEQv}
(\sum_{d \in \mathcal{D}}-\lambda_d)+ |\mathcal{D}|\min(0,\min_{p \in \mathcal{P}}\Gamma_p + \sum_{d \in \mathcal{D}}P_{dp}\lambda_{d})
\end{align}
Observe that the minimization in Eq \ref{LBEQv} is computed each time we do pricing in Alg \ref{dualSolve}.
\subsection{Pricing Using a Naive Dynamic Program}
\label{basicPricing}
Observe that Eq \ref{PricingEq} corresponds to computing the maximum a posteriori probability (MAP) of a Markov random field (MRF) where there is  bijection between parts (except the neck) and variables in the MRF.  Similarly for a variable in the MRF there is a bijection between the state space of that variable and the power set of detections for the associated part.  This MRF is tree structured and hence amenable to exact inference via dynamic programming which we now consider.  

We use $\mathcal{S}^r$ to denote the state space of variable $r$ which we describe using $S^r\in \{ 0,1\}^{|\mathcal{D}^{r}| \times 2^{|\mathcal{D}^r|} }$.  We index $S^{r}$ with $d,s$ where $d\in \mathcal{D}^r$ and $s \in \mathcal{S}^r$ respectively.  We  use $S^r_{ds}=1$ to indicate that detection $d\in \mathcal{D}^r$ is included in configuration $s \in \mathcal{S}^r$. We use $\mu_{\hat{r}\hat{s}\leftarrow r}$ to refer to the value of the lowest cost solution to the sub-tree rooted at $r$ conditioned on its parent $\hat{r}$ taking on state $\hat{s}$.  The cost $\mu_{\hat{r}\hat{s}\leftarrow r}$ includes the pairwise interaction terms between members of $\mathcal{D}^{r}$ and $\mathcal{D}^{\hat{r}}$. 
We  define $\mu$ formally using helper terms $\Phi,\psi$ below. 
\begin{align}
\mu_{\hat{r}\hat{s} \leftarrow r}=\min_{s \in \mathcal{S}^r}\Phi_{\hat{s}s}+\psi_{rs}+\sum_{\bar{r} \in T_{r\bar{r}=1}}\mu_{rs \leftarrow \bar{r}}\\
\nonumber \Phi_{\hat{s}s}= \sum_{\substack {d_1 \in \mathcal{D}^{\hat{r}}\\ d_2 \in \mathcal{D}^{r}}}\theta^2_{d_1d_2}S^{\hat{r}}_{d_1\hat{s}}S^r_{d_2s}\\
\nonumber \psi_{rs}=\sum_{d \in \mathcal{D}^r}(\theta^1_d+\lambda_d)S^r_{ds}+\sum_{\substack {d_1\in \mathcal{D}^r \\ d_2 \in \mathcal{D}^r}}\theta^2_{d_1d_2}S^{r}_{d_1s}S^r_{d_2s}
\\ \nonumber +\sum_{\substack{d_1 \in \breve{\mathcal{D}} \\ d_2 \in \mathcal{D}^{r }}}\theta^2_{d_1d_2}S^r_{d_2s}
\end{align}
Computing $\mu_{\hat{r}\hat{s} \leftarrow r} $ for all $\hat{s} \in \mathcal{S}^{\hat{r}}$ is expensive since we need to search over all possible combinations of state spaces of two adjacent variables, where the number of possible states of each variable can be up to 50k in our experiments.
\section{A Nested Benders Decomposition Alternative to Naive Dynamic Programming}
\label{bendPricing}
We now consider  NBD as an alternative  to naive dynamic programming that avoids the expensive computation of  $\mu_{\hat{r}\hat{s} \leftarrow r} $ for all $\hat{s} \in \mathcal{S}^{\hat{r}}$.  We express $\mu$ using the sum of  convex functions  each constructed from the maximum of a unique set of affine functions called Benders rows.  Specifically for any part $r\in \mathcal{R}$ (other than the root $T_0$) we denote the corresponding set of Benders rows  as $\mathcal{Z}^r$ which we index by $z$. We describe $\mathcal{Z}^r$ using $\Omega^r \in \mathbb{R}^{ |\mathcal{Z}^r|\times (1+|\mathcal{D}^{\hat{r}}|)}$ where $\hat{r}$ is the parent of $r$ and index with $(z,d)$ or $(z,0)$.  For a given $z$ we use $\Omega^r_{zd}$ to indicate value associated with $d \in \mathcal{D}^{\hat{r}}$ and $\Omega^r_{z0}$ to be  the offset.  Using $\mathcal{Z}$ we provide an alternative description of  $\mu$ using helper term $\mu^*$ defined as follows.  
\begin{align}
\label{minurs}
&\mu_{\hat{r}\hat{s}\leftarrow r}=\min_{s \in \mathcal{S}^{r}}\Phi_{\hat{s}s}+\mu^*_{rs}\\
\nonumber &\mu^*_{rs}=\psi_{rs}+\sum_{\substack{\bar{r} \in \mathcal{R}\\ T_{r\bar{r}}=1}}\max_{z \in \mathcal{Z}^{\bar{r}}}(\Omega^{\bar{r}}_{z0}+\sum_{d \in \mathcal{D}^r}\Omega^{\bar{r}}_{zd}S^{r}_{ds})
\end{align}
The existence of such a decomposition is a known result of the stochastic programming literature \cite{birge1985decomposition}.  Notice that the minimization in Eq \ref{minurs} does not require either the configuration of the children of $r$ nor ``messages" from those children.  Hence the state of $T_0$ can be determined independently of the other variables.   Similarly each variable can be determined independently of its children given the state of its parent.  If the sets $\mathcal{Z}^r$ are of small cardinality then solving Eq \ref{minurs} is easy thus we construct  a sufficient subset of $\mathcal{Z}^r$ denoted $\ddot{\mathcal{Z}}^r$ for each $r$ (other than the root) using row generation (cutting planes).  We refer to the collection of the nascent sets as $\ddot{\mathcal{Z}}$. Given the nascent sets $\ddot{\mathcal{Z}}$ we construct a lower bound on $\mu_{\hat{r}\hat{s} \leftarrow r}$ denoted $\mu^-_{\hat{r}\hat{s} \leftarrow r}$ defined below using helper function $\mu^{*-}_{rs}$.   
\begin{align}
&\mu^-_{\hat{r}\hat{s}\leftarrow r}=\min_{s \in \mathcal{S}^{r}}\Phi_{\hat{s}s}+\mu^{*-}_{rs}\\
\nonumber &\mu^{*-}_{rs}=\psi_{rs}+\sum_{\substack{\bar{r} \in \mathcal{R}\\ T_{r\bar{r}}=1}}\max_{z \in \ddot{\mathcal{Z}}^{\bar{r}}}(\Omega^{\bar{r}}_{z0}+\sum_{d \in \mathcal{D}^r}\Omega^{\bar{r}}_{zd}S^r_{ds})
\end{align}
The root variable $T_0$ is not associated with  $\mu^-_{\hat{r}\hat{s}\leftarrow r}$ terms since it has no parent but it is associated with $\mu^{*-}_{rs}$ terms.
\subsection{Overview of Constructing $\ddot{\mathcal{Z}}$ }
We now consider the construction of small sufficient sets $\ddot{\mathcal{Z}}$ such that $\min_{s \in S^{T_0}}\mu^{*-}_{T_0s}=\min_{s \in S^{T_0}}\mu^{*}_{T_0s}$.  We outline the remainder of this section as follows.  In Section \ref{ubUpDown} we produce upper/lower bounds on the MAP of the MRF, which are identical at termination of NBD.  The upper bound is accompanied by a configuration with cost equal to the upper bound.  In Section \ref{SelectNode} we compute the gap between the upper and lower bounds introduced at each variable in the tree and select the variable $r_*$ associated with the largest increase in the gap.  Then in Section \ref{genRows} we add a Benders row to $\ddot{\mathcal{Z}}^{r_*}$.  Next in Section \ref{reUse} we increase $\Omega^r_{0z}$ terms hence tightening the relaxation without generating new rows.  
  Then in Section \ref{AlgNBDesc} we combine the steps above to produce a complete NBD inference approach.  Finally we provide implementation details in Section \ref{impDetails}.
\subsection{Producing a Configuration and Corresponding  Bounds  using Nested Benders Decomposition}
\label{ubUpDown}
In this section produce a configuration for the MRF by proceeding from root to leaves and selecting the state for a given variable $r$ given the state of its parent only.   We describe the configuration produced using $x$ where $x_r \in S^{r}$. We use $x_{\hat{r}}$ to refer to the state of the parent of $r$.  The process of producing $x$ is defined below.
\begin{align}
&x_{T_0} \leftarrow \mbox{arg}\min_{s\in \mathcal{S}^{T_0}}\mu^{*-}_{T_0s}\\
\nonumber &x_{r} \leftarrow \mbox{arg}\min_{s\in \mathcal{S}^r}\Phi_{x_{\hat{r}}s}+\mu^{*-}_{rs}\quad \forall r \in \mathcal{R}-T_0
\end{align}
The cost of the configuration is an upper bound on the MAP and is associated with a lower bound on the MAP $ \min_{s \in \mathcal{S}^{T_0}}\mu^{*-}_{T_0s}$ .
\subsection{Computing the Gap Introduced at each Variable in the Tree}
\label{SelectNode}
In this section we identify the variable in the tree associated with the largest increase in the gap between the upper and lower bounds given a configuration $x$ produced in Section \ref{ubUpDown}.  The gap between upper and lower bounds introduced at $r$ is the difference between the upper and lower bounds at variable $r$ minus the corresponding gaps at its children. We use $M^1_r,M^2_r,M^3_r$, which are defined below, to denote the cost of the sub-tree rooted at $r$; the corresponding lower bound; and the gap introduced at $r$ respectively.  
\begin{align}
&M^1_{r}= \Phi_{x_{\hat{r}}x_{r}}+\psi_{x_{\hat{r}}x_r}+\sum_{\bar{r} \in T_{r,\bar{r}}=1} M^1_{\bar{r}}\\
\nonumber  &M^2_{r}=\max_{z \in \ddot{\mathcal{Z}}^{r}}(\Omega^r_{z0}+\sum_{d \in \mathcal{D}^{\hat{r}}}\Omega^r_{zd}S^{\hat{r}}_{dx_{\hat{r}}})\\
\nonumber &M^3_r=(M^1_{r}-M^2_{r})-\sum_{\bar{r} \in T_{r,\bar{r}}=1}(M^1_{\bar{r}}-M^2_{\bar{r}})
 \end{align}
\subsection{Generating Benders Rows}
\label{genRows}
In this section we identify the most violated Benders row denoted $z$ for a given $r$, given that its parent $\hat{r}$ takes on $\hat{s}$ and the $\ddot{\mathcal{Z}}$ sets.  We formulate this as a small scale linear program described below.

We use decision variable $[x_{r}=s]=1$ to indicate that $s$ is the state associated with variable $r$.  We use decision variable $y \in \mathbb{R}_{0+}^{|\mathcal{D}^{\hat{r}}|\times|\mathcal{D}^{r}|}$ which we index by $d_1,d_2$ where $y_{d_1d_2}=S^{\hat{r}}_{d_1\hat{s}}(\sum_{s \in \mathcal{S}^r}[x_{r}=s]S^r_{d_2s})$. We introduce dual variables $\delta^{0z} \in \mathbb{R}$ and $\delta^{1z},\delta^{2z},\delta^{3z}$ each of which lie in $\mathbb{R}_{0+}^{|\mathcal{D}^r|}$.
\begin{align}
\label{primalFormGetRow}
\min_{s \in S^r} \Phi_{\hat{s}s}+\mu^{*-}_{rs}\\
\nonumber =\min_{\substack{\sum_{s \in \mathcal{S}^r} [x_{r}=s] = 1 \\   -y_{d_1d_2}+S^{\hat{r}}_{d_1\hat{s}}+ \sum_{s\in \mathcal{S}^r} [x_{r}=s]S^r_{d_2s}\leq 1\\ y_{d_1d_2}\leq S^{\hat{r}}_{d_1\hat{s}} \\ y_{d_1d_2}\leq \sum_{s\in \mathcal{S}^r} [x_{r}=s] S^r_{d_2s}\\ x \geq 0 \\ y \geq 0}} \sum_{s\in \mathcal{S}^r} [x_{r}=s]\mu^{*-}_{rs}\\ 
\nonumber+\sum_{\substack{d_1 \in \mathcal{D}^{\hat{r}} \\ d_2 \in \mathcal{D}^r}}\theta^2_{d_1d_2}y_{d_1d_2}\\
\nonumber = \max_{\substack{\delta^{0z} \in \mathbb{R} \\ \delta^{1z}_{d_1d_2}\geq 0 \\ \delta^{2z}_{d_1d_2}\geq 0\\ \delta^{3z}_{d_1d_2}\geq 0 }}
-\delta^{0z} + \sum_{\substack{d_1 \in \mathcal{D}^{\hat{r}} \\ d_2 \in \mathcal{D}^r}}\delta^{1z}_{d_1d_2}(S^{\hat{r}}_{d_1\hat{s}}-1)-\delta^{2z}_{d_1d_2}S^{\hat{r}}_{d_1\hat{s}} \\
\nonumber \mbox{s.t. }\mu^{*-}_{rs}+\delta^{0z}+\sum_{\substack{d_1 \in \mathcal{D}^{\hat{r}} \\ d_2 \in \mathcal{D}^r}} (\delta^{1z}_{d_1d_2}-\delta^{3z}_{d_1d_2})S^r_{d_2s} \geq 0 \quad \forall s \in \mathcal{S}^r\\
\nonumber \theta^2_{d_1d_2}-\delta^{1z}_{d_1d_2}+\delta^{2z}_{d_1d_2}+\delta^{3z}_{d_1d_2}\geq 0 \quad \forall ( d_1 \in \mathcal{D}^{\hat{r}},d_2\in \mathcal{D}^r)
\end{align}
After computing $\delta^{0z},\delta^{1z},\delta^{2z},\delta^{3z}$ we produce a new Benders row denoted $z$ that is defined as follows.  
\begin{align}
\Omega^r_{z0}\leftarrow -\delta^{0z}-\sum_{\substack{d_1 \in \mathcal{D}^{\hat{r}} \\ d_2 \in \mathcal{D}^r}}\delta^{1z}_{d_1d_2}\\
\nonumber \Omega^r_{zd_1}\leftarrow \sum_{d_2 \in \mathcal{D}^r }\delta^{1z}_{d_1d_2}-\delta^{2z}_{d_1d_2} \quad d_1 \in \mathcal{D}^{\hat{r}} 
\end{align}
When solving optimization in the dual of  Eq \ref{primalFormGetRow} we add a tiny negative bias to objective corresponding to terms $\delta^{1z},\delta^{2z},\delta^{3z}$ . This ensures that the corresponding terms do not increase beyond what is needed to produce an optimal dual solution, which stabilizes optimization.
The additional small biases may be understood intuitively as ensuring that $\sum_{d_2 \in \mathcal{D}^r}\delta^1_{d_1d_2}-\delta^2_{d_1d_2}$ corresponds to the marginal cost for using $d_1$ in the solution. In Appendix \ref{CompressedNB} we solve the dual LP in Eq \ref{primalFormGetRow} efficiently by reducing it to an equivalent LP with $|\mathcal{D}^r|$ variables and far fewer constraints.
\subsection{Re-using Rows: Rapidly Updating $\delta^{0z}$ while leaving $\delta^{1z},\delta^{2z},\delta^{3z}$ fixed }
\label{reUse}
In this sub-section we provide a complementary mechanism to generating new Benders rows.   This mechanism is motivated by the observation that $\mu^{*-}_{rs}$ may increase (but never decrease) when Benders rows are added to the descendants of a given variable $r$. This mechanism takes in existing Benders rows and sets the corresponding  $\delta^{0z}$ term to the minimum feasible value thus tightening the corresponding constraint while leaving $\delta^{1z},\delta^{2z},\delta^{3z}$ fixed.  This task is faster than generating Benders rows via the method of Section \ref{genRows}.

Observe that given $\delta^{1z},\delta^{2z},\delta^{3z}$ satisfying $\theta^2_{d_1d_2}-\delta^{1z}_{d_1d_2}+\delta^{2z}_{d_1d_2}+\delta^{3z}_{d_1d_2}\geq 0$ there always exists a feasible setting of $\delta^{0z}$.  
Given fixed $\delta^{1z},\delta^{2z},\delta^{3z}$ we select the smallest feasible value for $\delta^{0z}$ as follows.  
\begin{align}
\label{tightEq}
\delta^{0z}\leftarrow -(\min_{s \in \mathcal{S}^r}\mu^{*-}_{rs}+\sum_{\substack{d_1 \in \mathcal{D}^{\hat{r}} \\ d_2 \in \mathcal{D}^r}} (\delta^{1z}_{d_1d_2}-\delta^{3z}_{d_1d_2})S^r_{d_2s})
\end{align}
%
%
\subsection{Our Complete Nested Benders Decomposition Algorithm}
\label{AlgNBDesc}
Our NBD approach iterates through the following steps.  
\begin{enumerate}
\item
\textbf{Step 1}:   Proceeding from the leaves to the children of the root:   Set $\delta^{0z}$ terms to the minimum feasible value given fixed $\delta^{1z},\delta^{2z},\delta^{3z}$.  This is done on the first iteration of NBD only if $\ddot{\mathcal{Z}}^r$ is not empty for each $r$.
\item \textbf{Step Two}:  Proceed from root to leaves: select the state for each variable  conditioned on its parent (if it has one) and the Benders rows associated with its children.  This produces upper and lower bounds associated with each variable in the tree. 
\item  \textbf{Step Three}:   Select the variable  $r_*$ corresponding to the largest increase in the gap between the upper and lower bounds in the tree.
\item  \textbf{Step Four}:  Add a new Benders row to $\ddot{\mathcal{Z}}^{r_*}$ 
\end{enumerate}
We repeat this procedure until no additional Benders rows need be added at which point the configuration produced in step two is guaranteed to be the global optima. We formalize this procedure in Alg \ref{main2}. 
\begin{figure}
 \begin{algorithm}[H]
 \caption{Overall Algorithm of Nested Benders Given Initial $\ddot{\mathcal{Z}}$}
\begin{algorithmic} 
\Repeat
\State \textbf{Step 1:  Update $\delta^0$ terms}
\For{$r \in \mathcal{R}$ from leaves to children of the root}
\For{$z \in \ddot{\mathcal{Z}}^r$ } 
\State $\delta^{0z}\leftarrow -\min_{s\in \mathcal{S}^r}\mu^{*-}_{rs}+\sum_{\substack{d_1 \in \mathcal{D}^{\hat{r}} \\ d_2 \in \mathcal{D}^r}} (\delta^{1z}_{d_1d_2}-\delta^{3z}_{d_1d_2})S^r_{d_2s}$
\State $\Omega^r_{z0}\leftarrow -\delta^{0z}-\sum_{d \in \mathcal{D}^r}\delta^{1z}_{d}$
\EndFor
\EndFor
\State \textbf{STEP 2: Compute configuration and bounds}
\State $x_{T_0} \leftarrow \mbox{arg}\min_{s\in \mathcal{S}^{T_0}}\mu^{*-}_{T_0s}$%
\For{$r \in \mathcal{R}$ \quad from children of  $T_0$  to leaves} 
\State $x_{r} \leftarrow \mbox{arg}\min_{s\in \mathcal{S}^r}\Phi_{x_{\hat{r}}s}+\mu^{*-}_{rs}$
\EndFor
\State $p\leftarrow$ pose corresponding to $x$
\State $UB \leftarrow \Gamma_p+\sum_{d \in \mathcal{D}}\lambda_d P_{dp}$ 
\State $LB \leftarrow \min_{s\in \mathcal{S}^{T_0}}\mu^{*-}_{T_0s}$
\State \textbf{STEP 3:  Select variable to add Benders row to }
\For{$r \in \mathcal{R}$ from leaves to children of $T_0$} 
\State$ M^1_{r}\leftarrow \Phi_{x_{\hat{r}}x_{r}}+\psi_{x_{\hat{r}}x_r}+\sum_{\bar{r} \in T_{r,\bar{r}}=1} M^1_{\bar{r}}$
\State $M^2_{r}=\max_{z \in \ddot{\mathcal{Z}}^{r}}(\Omega^r_{z0}+\sum_{d \in \mathcal{D}^{\hat{r}}}\Omega^r_{zd}S^{\hat{r}}_{dx_{\hat{r}}})$
\State $M^3_r=(M^1_{r}-M^2_{r})-\sum_{\bar{r} \in T_{r,\bar{r}}=1}(M^1_{\bar{r}}-M^2_{\bar{r}})$
\EndFor
\State $r_* \leftarrow \mbox{arg}\max_{r \in \mathcal{R}-T_0}M^3_r$
\State \textbf{STEP 4:  Generate a new Benders row for $r_*$}
\If {$UB\neq LB$}
\State generate $\delta^{0z},\delta^{1z},\delta^{2z},\delta^{3z}$ by solving dual in Eq \ref{primalFormGetRow} given parent taking on state $x_{\hat{r}_*}$.
\State $\Omega^{r_*}_{z0}\leftarrow -\delta^{0z}-\sum_{\substack{d_1 \in \mathcal{D}^{\hat{r}} \\ d_2 \in \mathcal{D}^r}}\delta^{1z}_{d_1d_2}$\\
\State $ \Omega^{r_*}_{zd_1}\leftarrow \sum_{d_2 \in \mathcal{D}^r }\delta^{1z}_{d_1d_2}-\delta^{2z}_{d_1d_2} \quad d_1 \in \mathcal{D}^{\hat{r}}$
\State $\ddot{\mathcal{Z}}^{r_*} \leftarrow \ddot{\mathcal{Z}}^{r_*}\cup z $
\EndIf
 \Until{ \mbox{UB}=\mbox{LB}}
\end{algorithmic}
\label{main2}
  \end{algorithm}
\end{figure}
\subsection{Implementation Details }
\label{impDetails}
\begin{figure*}
\begin{center}
  \begin{tabular}{c c} 
      \includegraphics[clip,width=.45\textwidth]{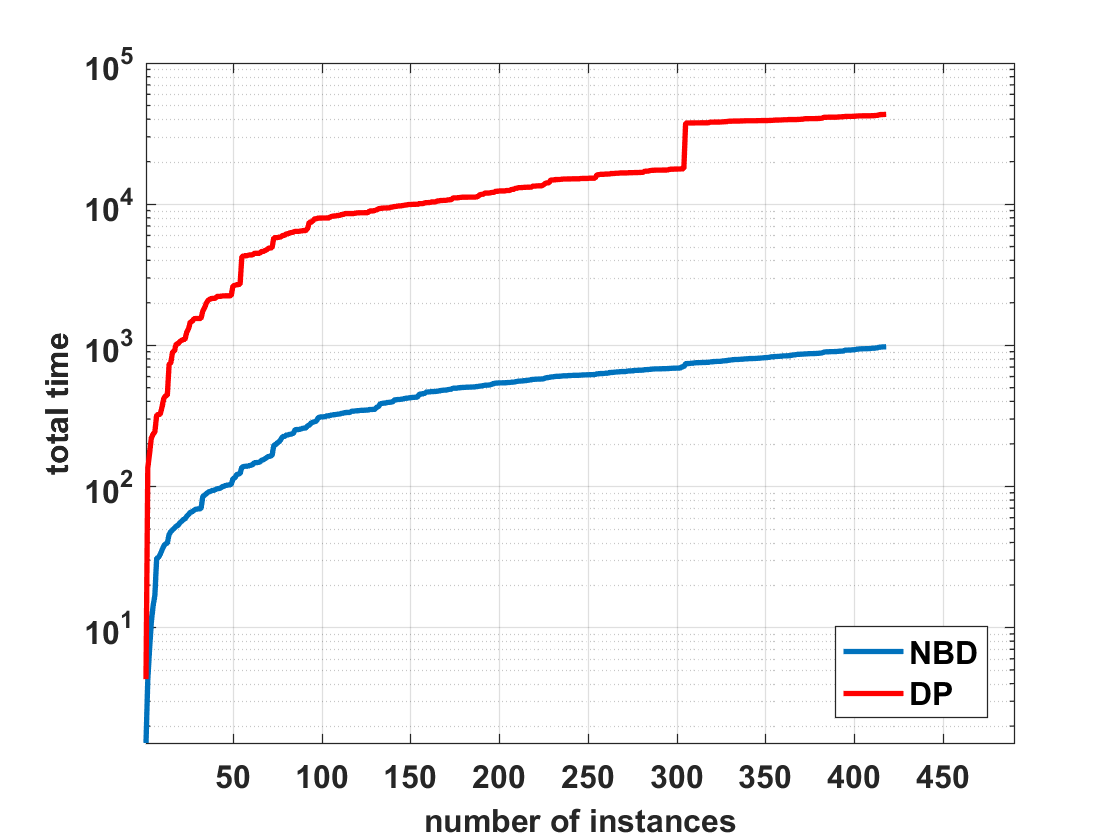} &
      \includegraphics[clip,width=.45\textwidth]{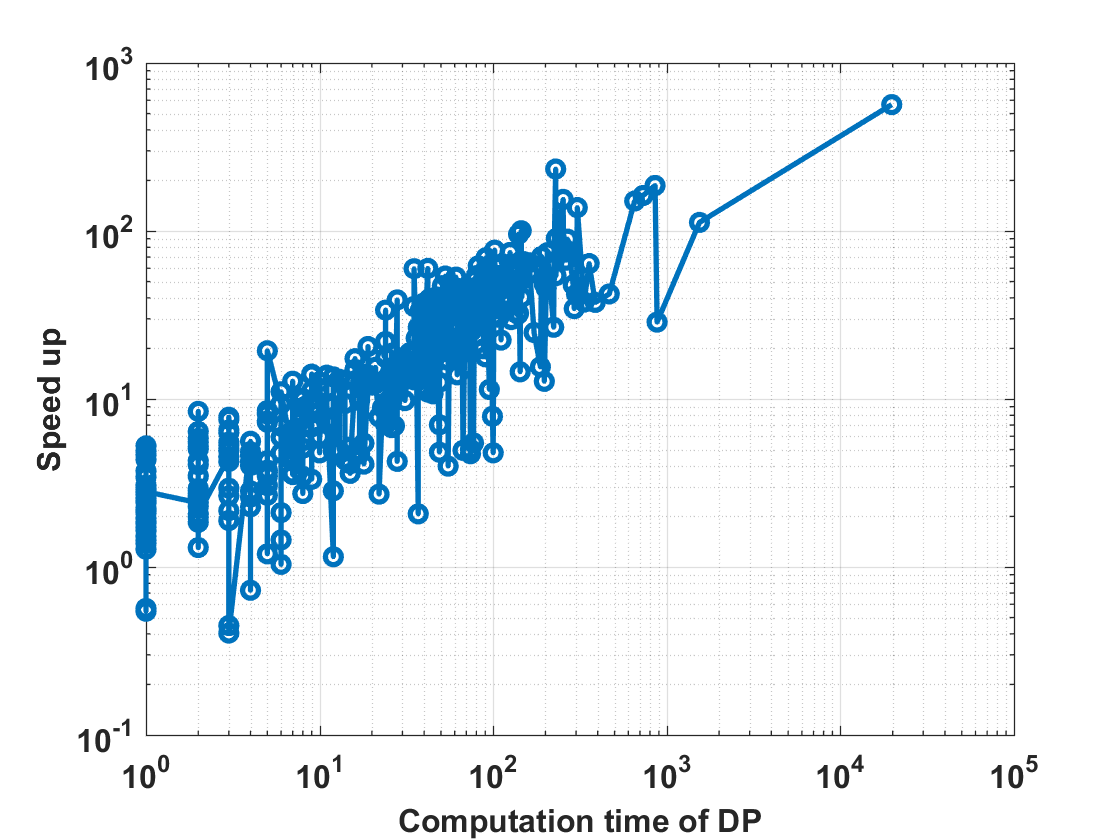}\\
      (a) & (b)
  \end{tabular}
\end{center}
    \caption{Timing comparison and speed-ups achieved by NBD. (a) Accumulated running time over problem instances for NBD and DP, respectively. (b) Factor of speed-up of NBD relative to DP, as a function of computation time spent for DP pricing. Note that in general the factor of speed-up grows as the problem gets harder for DP.}
  \label{fig:timing}
\end{figure*}
In this section we provide implementation details with regards Alg \ref{main2}.  

\textbf{Accelerating Step One of Alg \ref{main2}:}  Observe that  $\delta^{0z}$ terms associated with $\ddot{\mathcal{Z}}^r$  can only be decreased if a new Benders row is added to one of the descendants of $r$.  Thus when executing Alg \ref{main2} we only update the $\delta^0$ terms associated with the ancestors of the variable  $r_*$ which had its Benders row set augmented.  Observe that we need only update the $\delta^{0z}$ terms associated with the leaves in the first iteration of NBD in a given call from ICG.

\textbf{Accelerating Step Two of Alg \ref{main2}}.   Recall that the choice of the state of a given variable $r$ is a function of the Benders rows associated with its children and the configuration of the parent (if it has a parent).  At any iteration of NBD other than the first we consider the  previous configuration of the MRF produced in step two of Alg \ref{main2} and only update the state of a variable  if the state of its parent was changed or if it is an ancestor of the variable $r_*$ which had its Benders row set augmented.

\textbf{Limiting the Number of Neck Detections in a Pose:  }
We found that our best results with regards to timing and modeling occur when we require that each pose include exactly one neck detection.

\textbf{Limiting State Space of a Variable:  }
We limit the number of states of a given variable to a given user defined parameter value $V$ ($V$=50,000).  We construct this set as follows.  We begin with the state corresponding to zero detections included, then add in the group of  states corresponding to one detection included;  then add in the group of  states corresponding to two detections included etc.  If adding a group would have the state space exceed $V$ states for the variable we don't add the group and terminate.  

\textbf{Caching Integrals:  }
We accelerate Alg \ref{main2} by storing the value of  repeatedly used integrals that do not change in value over the course of optimization.

Thus each time a new $z$ is produced we store $\sum_{d \in \mathcal{D}^{\hat{r}}} \Omega^{r}_{zd}S^{\hat{r}}_{d\hat{s}}$ for each $\hat{s} \in \mathcal{S}^{\hat{r}}$.  Similarly  we store $\sum_{\substack{d_1 \in \mathcal{D}^{\hat{r}} \\ d_2 \in \mathcal{D}^r}}(\delta^{1z}_{d_1d_2}-\delta^{3z}_{d_1d_2})S^r_{d_2s}$ for each $s \in \mathcal{S}^r$.

\textbf{Initializing $\ddot{\mathcal{Z}}$:  }
We do not initialize $\ddot{\mathcal{Z}}^r$ with any Benders rows for the first round of pricing in ICG.  Thus the initial state for a variable in NBD ignores the existence of its children and the corresponding  initial lower bound is $-\infty$.

\textbf{Timing Observation:  }
Experimentally we observe that the total time consumed by steps in NBD is ordered from greatest to least as [1,2,4,3].  Note that the step solving the LP is the second least time consuming step of NBD.

\textbf{Selecting the Root:  }
 Observe that Alg \ref{main2} requires solving LPs in step four for variables except the root.  The number of constraints in the  LP for part $r$ is exponential in the size of $|\mathcal{D}^r|$.  We avoid solving the largest LP by selecting as the root the part associated with the most detections. Alternatively we could select the root  so as to have a more balanced tree with the goal of leveraging parallel processing.  
\section{Experiments}
\label{exper}
\begin{table*}
\begin{center}
\resizebox{.95\textwidth}{!}{
\renewcommand{\arraystretch}{1.5}
 \begin{tabular}{||c c c c c c c c c c | c||} 
 \hline
 Part & Head & Shoulder & Elbow & Wrist  & Hip & Knee& Ankle & mAP(UBody) & mAP & time (s/frame)\\ [0.5ex] 
 \hline\hline
 \textbf{Ours} & 90.6  & 87.3  & 79.5  & 70.1  & 78.5  & 70.5 & 64.8 & 81.8 & 77.6 & 1.95 \\ 
 \hline
 \cite{NL-LMP} & 93.0  & 88.2  & 78.2  & 68.4  & 78.9  & 70.0 & 64.3 & 81.9 & 77.6 & 0.136 \\ \hline
\end{tabular}
}
\caption{We display average precision of our approach versus~\cite{NL-LMP}. Running times are measured on an Intel i7-6700k quad-core CPU.}
 \label{tab:mAP}
\end{center}
\end{table*}
\begin{figure*}
\begin{center}
  \begin{tabular}{c c c c} 
      \includegraphics[clip,trim=.5cm 2cm 1cm 2cm,width=.2\textwidth]{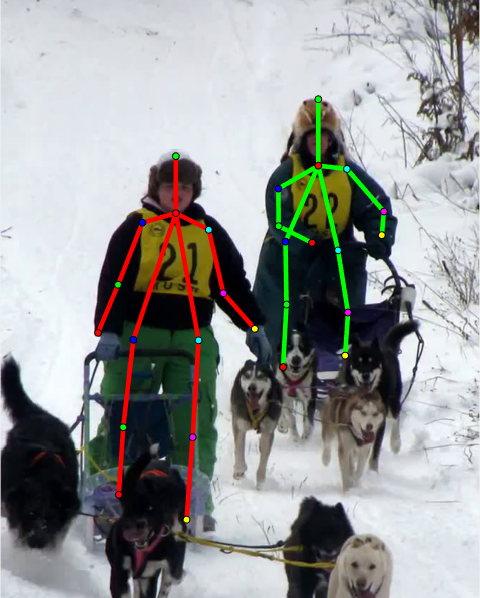} &
      \includegraphics[clip,trim= 2cm 0cm 2cm 0cm,width=.255\textwidth]{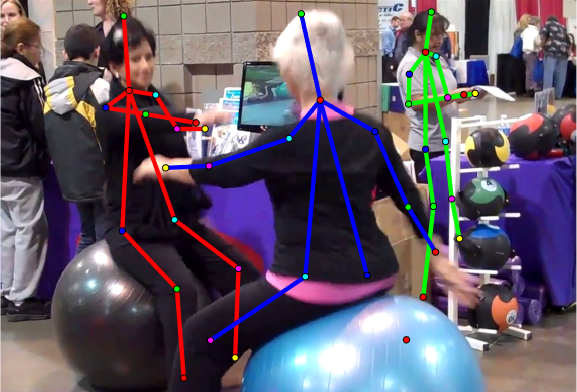} &
      \includegraphics[clip,trim=.5cm 1cm 1cm 1cm,width=.16\textwidth]{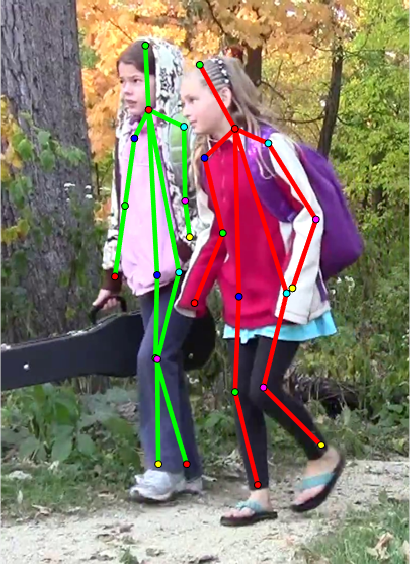} &
      \includegraphics[clip,trim=3cm 2cm 1cm 1cm,width=.2\textwidth]{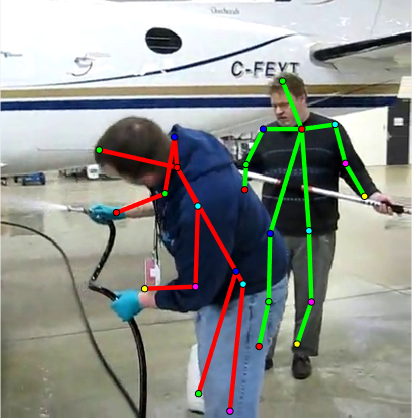}\\
      \multicolumn{2}{c}{
      \includegraphics[clip,trim=0cm 0cm 3cm 1cm,width=.4\textwidth]{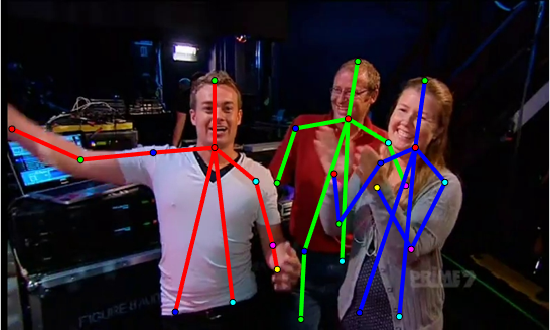}
      } &
      \includegraphics[clip,trim= 2cm 2cm 1cm 2cm,width=.18\textwidth]{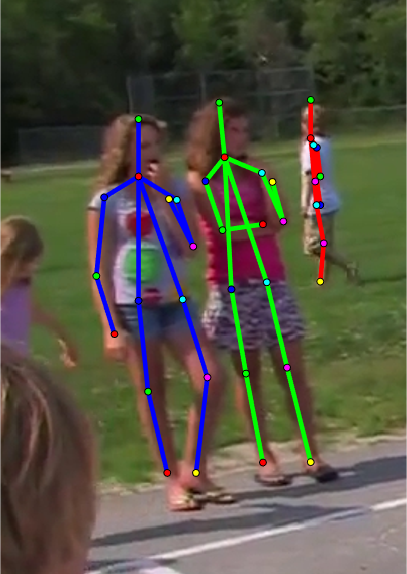} &
      \includegraphics[clip,trim=1cm 1cm 2cm 1cm,width=.205\textwidth]{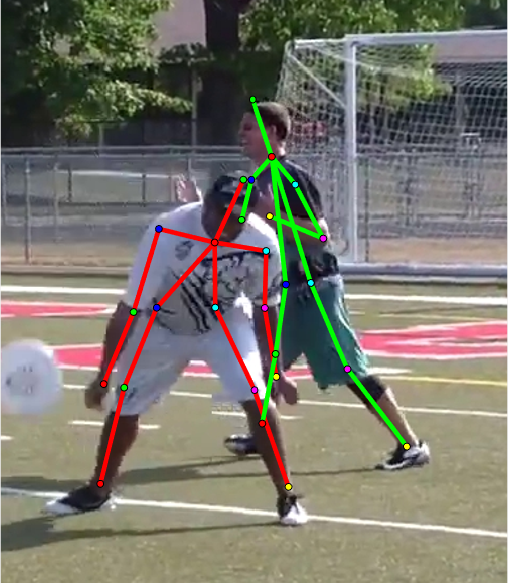}
  \end{tabular}
\end{center}
    \caption{Example output of our system.}
  \label{fig:vis}
\end{figure*}
We evaluate our approach against a naive dynamic programming based formulation on MPII-Multiperson validation set~\cite{mpiiBenchmark}, which consists of 418 images. The unary $\theta^1$ and pairwise $\theta^2$ costs are trained using the code of~\cite{deepcut2}, with a constant bias $\theta^0=60$ (set by hand) to regularize the number of people in the solution.

We compare solutions found by NBD and DP at each step of ICG; for all problem instances and all optimization steps, NBD obtains exactly the same solutions as DP (up to a tie in costs). Comparing total time spent doing NBD vs DP across problem instances we found that NBD is 44x faster than DP, and can be up to 500x faster on extreme problem instances. Comparison of accumulated running time used by NBD and DP over all 418 instances are shown in Fig.~\ref{fig:timing}.   We observe that the factor speed up provided by NBD increases as a function of the computation time of DP.   

With regards to cost we observe that the integer solution produced over $\hat{\mathcal{P}}$ is identical to the LP value in over 99\% of problem instances thus certifying that the optimal integer solution is produced.  For those instances on which LP relaxation fails to produce integer results, the gaps between the LP objectives and the integer solutions are all within 1.5\% of the LP objectives. 

For the sake of completeness, we also report MPPE accuracy in terms of average precisions (APs) and compare it against a state-of-the-art solver~\cite{NL-LMP} which uses primal heuristics. Note that the cost formulation of~\cite{NL-LMP} differs from ours in that it allows a  pose to be associated with multiple neck detections or none, while our model requires that each pose must have exactly one neck detection and maps detections to parts prior to ICG.  Also our model includes a prior on the number of poses as modeled by $\theta^0$. As shown in Table~\ref{tab:mAP}, we achieve equivalent results to~\cite{NL-LMP}. We note that although our algorithm does not run as fast as~\cite{NL-LMP}, our code is implemented in pure MATLAB and can benefit further from using commercial LP solvers and parallelizing pricing routines. More importantly, our formulation provides upper/lower bounds and in over 99\% of cases certificates of optimality.
\section{Conclusion}
\label{conc}
We have described multi-person pose estimation as a minimum-weight set packing (MWSP) problem which we address using implicit column generation. We solve the corresponding pricing problem using a novel nested Benders decomposition (NBD) approach, which reuses Bender's rows between calls to NBD. For over 99\% of cases we find provably optimal solutions, which is practically important in domains where knowledge of certainty matters, such as interventions in rehabilitation. Our procedure for solving the pricing problem vastly outperforms a baseline dynamic programming approach.  We expect that NBD will find many applications in machine learning and computer vision, especially for solving dynamic programs with large state spaces for individual variables.  For example we could formulate sub-graph multi-cut tracking \cite{trackerbio} as a MWSP with pricing using NBD.  
{\small
\bibliography{fin_bib}
\bibliographystyle{ieee}
}
\pagebreak
\appendix
\section{Appendix:  Dual Optimal Inequalities on $\lambda$}
\label{dualBound}
In this section we provide upper bounds on the Lagrange multipliers $\lambda$ called dual optimal inequalities (DOI) \cite{ben2006dual} which are computed prior to ICG.  The use of DOI decreases the search space that ICG needs to explore and thus decreases the number of iterations of pricing required.  

Observe that at any given iteration of ICG the optimal solution to the primal LP relaxation need not lie in the span of  $\hat{\mathcal{P}}$.  If limited to producing a primal solution over $\hat{\mathcal{P}}$ it is useful to allow some values of $P\gamma$ exceed one.
We introduce a slack vector $ \xi \in \mathbb{R}_{0+}^{|\mathcal{D}|}$ indexed by $d$  that tracks the presence of any detections included more than once and prevents them from contributing to the objective when the corresponding contribution is negative. To do this we offset the cost for ``over-including" a detection with a cost that at least compensates and likely overcompensates.

Observe that the removal of a detection $d$ from a pose removes from the cost the associated  $\theta^1_d$, $\theta^2_{dd_1},\theta^2_{d_1d}$ for any $d_1$ in the pose and if $d$ is the only detection the $\theta^0$ term.  We define $\Xi_d$ such that it is an upper bound on the increase in the cost of a pose given that $d$ is removed.  To express $\Xi_d$ compactly we introduce the following terms  $\theta^{2+}_{d_1d_2}=\max(0,\theta^{2}_{d_1d_2})$,$\theta^{2-}_{d_1d_2}=\min(0,\theta^{2}_{d_1d_2})$,$\theta^{0-}=\min(0,\theta^{0})$.  We define $\Xi_d$ as follows.
\begin{align}
\Xi_d=-\min(0,\theta^{0-}+\theta^1_{d}+\sum_{d_1 \in \mathcal{D}}\theta^{2-}_{dd_1}+\theta^{2-}_{d_1d})
\end{align}
The expanded MWSP objective and its dual LP relaxation are given below:
\begin{align}
    \min_{\substack {\gamma_p \in \{0,1\} \\ \xi \geq 0\\ \sum_{p \in \mathcal{P}}P_{dp}\gamma_p-\xi_d \leq 1  \quad \forall d \in \mathcal{D}}} \sum_{p \in \mathcal{P}} \Gamma_p\gamma_p+\sum_{d \in \mathcal{D}}\Xi_d\xi_d \\
\nonumber =
\max_{\substack{\Xi \geq \lambda \geq 0 \\ \Gamma +P^{\top}\lambda \geq 0}}-1^{\top}\lambda
\end{align}
Observe that the dual relaxation bounds $\lambda$ by $\Xi$ from above.  These bounds are called DOI.

For cases where a pose is required to include a neck detection we can not use this bound for neck detections as the removal of the neck makes the pose invalid. Therefore we ignore the $\theta^{0-}$ term when computing $\Xi$ and for  $R_d=$ neck we set  $\Xi_{d}=\infty$. 

To ensure that the DOI are not active at termination of ICG we offset $\Xi$ with a small positive constant.  
\section{Appendix:  Deriving a Compressed LP for Benders Row Generation}
\label{CompressedNB}
In this Section we compress the dual form of the LP in Eq \ref{primalFormGetRow} so as to accelerate inference.  In fact by compressing the LP we observe that optimization of this LP no longer dominates NBD computation.  To achieve this we make the following observations about the primal LP form in Eq \ref{primalFormGetRow}.  
\begin{itemize}
\item 
Given $\theta^2_{d_1d_2}\leq 0$  or $S^{\hat{r}}_{d_1\hat{s}}=0$:  The constraint  $-y_{d_1d_2}+S^{\hat{r}}_{d_1\hat{s}}+ \sum_{s\in \mathcal{S}^r} [x_{r}=s]S^r_{d_2s}\leq 1$ is inactive so $\delta^{1z}_{d_1d_2}=0$.
\item
Given $\theta^2_{d_1d_2}\geq 0$ or $S^{\hat{r}}_{d_1\hat{s}}=1$: The constraint  $y_{d_1d_2}\leq S^{\hat{r}}_{d_1\hat{s}}$ is inactive so $\delta^{2z}_{d_1d_2}=0$.  
 \item 
Given $\theta^{2}_{d_1d_2}\geq 0$ the constraint $y_{d_1d_2}\leq \sum_{s\in \mathcal{S}^r} [x_{r}=s] S^r_{d_2s}$ inactive so $\delta^{3z}_{d_1d_2}=0$.
 \end{itemize}
Observe that the following is true  any pair $d_1 \in \mathcal{D}^{\hat{r}},d_2\in \mathcal{D}^{r}$ such that $\theta^2_{d_1d_2}<0$.
  \begin{align}
 \theta^2_{d_1d_2}+\delta^{2z}_{d_1d_2}+\delta^{3z}_{d_1d_2}\geq 0
 \end{align}
Since $\delta^{2z}_{d_1d_2}$ is non-negative and associated with a non-negative term in the objective we observe the following.
 \begin{align}
  \theta^2_{d_1d_2}+\delta^{2z}_{d_1d_2}=-\delta^{3z}_{d_1d_2}\\
\nonumber - \theta^2_{d_1d_2}\geq \delta^{2z}_{d_1d_2}
 \end{align}
 Using these observations we rewrite optimization.
 \begin{align}
 & \max_{\substack{ \delta^{0z} \in \mathbb{R} \\ \delta^{1z} \geq 0 \\ \delta^{2z}\geq 0\\S^{\hat{r}}_{d_1\hat{s}}\theta^{2+}_{d_1d_2}\geq  \delta^{1z}_{d_1d_2} \\  -(1-S^{\hat{r}}_{d_1\hat{s}})\theta^{2-}_{d_1d_2}\geq  \delta^{2z}_{d_1d_2} }}
\sum_{\substack{d_1 \in \mathcal{D}^{\hat{r}} \\ d_2 \in \mathcal{D}^r}}\delta^{1z}_{d_1d_2}(S^{\hat{r}}_{d_1\hat{s}}-1)-\delta^{2z}_{d_1d_2}S^{\hat{r}}_{d_1\hat{s}}-\delta^{0z} \\
\nonumber &\mbox{s.t. }\mu^{*-}_{rs}+\delta^{0z}+\sum_{\substack{d_1 \in \mathcal{D}^{\hat{r}} \\ d_2 \in \mathcal{D}^r}} (\delta^{1z}_{d_1d_2}+\delta^{2z}_{d_1d_2}+\theta^{2-}_{d_1d_2})S^r_{d_2s} \geq 0 
 \end{align}
Now observe that $\delta^{0z}$ is the only term that is associated with a non-zero objective and which is not bound to zero thus it has the same value as the primal LP and therefor $\delta^0=\min_{s \in S^r} \Phi_{\hat{s}s}+\mu^{*-}_{rs}$.  

Observe that for all $\delta^{1z},\delta^{2z}$ terms not bound to zero for a given $d_2$ that they co-occur in the objective with value zero and in each constraint over $S^r$ with the common value.  Using this we merge $\delta^{1z},\delta^{2z}$ terms across  $d_2$ as follows using $\delta^{4z} \in \mathbb{R}^{|\mathcal{D}^r|}$ which we  index by $d_2$ and helper term $Q_{d_2}$.  
 \begin{align}
 \delta^{4z}_{d_2}=\sum_{\substack{d_1\in \mathcal{D}^{\hat{r}}}}\delta^{1z}_{d_1d_2}+\delta^{2z}_{d_1d_2}\\
 \nonumber\delta^{1z}_{d_1d_2}=\delta^{4z}_{d_2}S^{\hat{r}}_{\hat{s}d_1}\frac{\theta^{2+}_{d_1d_2}}{Q_{d_2}}\\
 \nonumber \delta^{2z}_{d_1d_2}=\delta^{4z}_{d_2}(1-S^{\hat{r}}_{\hat{s}d_1})\frac{-\theta^{2-}_{d_1d_2}}{Q_{d_2}}\\
 \nonumber Q_{d_2}=\sum_{d_1 \in \mathcal{D}^{\hat{r}}} S^{\hat{r}}_{d_1\hat{s}}\theta^{2+}_{d_1d_2}-(1-S^{\hat{r}}_{\hat{s}d_1})\theta^{2-}_{d_1d_2}
 \end{align}
 We add a tiny magnitude negative objective to $\delta^{4z}_{d_2}$ for each $d_2$ so as to ensure that the smallest valued solution is produced.  
 This ensures that that the corresponding terms do not increase beyond what is needed to produce a dual feasible solution which stabilizes optimization.  We use $\epsilon=10^{-10}$ to express this.  
  \begin{align}
  \max_{\substack{ Q_{d_2}\geq \delta^{4z}_{d_2} \geq 0 }}
&-\epsilon \sum_{\substack{ d_2 \in \mathcal{D}^r}} \delta^{4z}_{d_2} \\
\nonumber &\mbox{s.t. }\mu^{*-}_{rs}+\delta^{0z}+\sum_{\substack{ d_2 \in \mathcal{D}^r}} (\delta^{4z}_{d_2}+\sum_{d_1 \in \mathcal{D}^{\hat{r}}}\theta^{2-}_{d_1d_2})S^r_{d_2s} \geq 0 
 \end{align}
We now remove  constraints in the LP corresponding to members of $S^r$(constraints of the form  $\mu^{*-}_{rs}+\delta^{0z}+\sum_{\substack{ d_2 \in \mathcal{D}^r}} (\delta^{4z}_{d_2}+\sum_{d_1  \in \mathcal{D}^{\hat{r}}}\theta^{2-}_{d_1d_2})S^r_{d_2s} \geq 0$) without altering the solution of the LP. 

Observe that the slack in these constraints does not decrease as $\delta^{4z}$ increases.  Thus to determine which constraints of the form  above that we need not consider we set $\delta^{4z}$ to be the zero vector.  Any such constraints that are are not violated will not be violated for any setting of $\delta^{4z}$ and so can be ignored when solving the LP.  In practice we find that that the proportion of the constraints that we remove is  very large and thus we achieve considerable time savings when solving the LP.  

\end{document}